\documentclass[conference]{IEEEtran}

\usepackage{booktabs}
\usepackage{array}
\usepackage{balance}
\usepackage{graphicx}
\usepackage{amsmath}
\usepackage{url}

\title{Toward Workflow-Aware Benchmarking for Healthcare NLP Agents}

\author{
\IEEEauthorblockN{Junyi Yao\IEEEauthorrefmark{1}, Baichuan Li\IEEEauthorrefmark{2}, Zihao Zheng\IEEEauthorrefmark{1}, Jiayu Long\IEEEauthorrefmark{1}}
\IEEEauthorblockA{
\IEEEauthorrefmark{1}Washington University in St. Louis, USA\\
\{j.yao,z.zihaogary,jiayujacqueline\}@wustl.edu}
\IEEEauthorblockA{
\IEEEauthorrefmark{2}Southern Methodist University, USA\\
baichuanl@smu.edu}
}

\begin{document}
\maketitle

\begin{abstract}
Large language model (LLM) agents are increasingly proposed for healthcare tasks such as clinical documentation, evidence retrieval, patient messaging, and care coordination. Yet many evaluations remain limited to static medical question answering or one-shot generation, under-representing longitudinal state, interruptions, and human handoffs. We introduce an episode-level evaluation protocol for healthcare NLP agents. The protocol separates evidence across model, agent, and simulated-workflow behavior; specifies a five-field episode schema; and defines annotation and scoring for state continuity, evidence traceability, and escalation decisions. It is instantiated as four task templates: documentation update, evidence retrieval, patient messaging, and triage handoff. The protocol does not claim to measure clinical outcomes or deployment value. Instead, it supplies a reproducible intermediate evaluation layer between static benchmarks and prospective workflow studies, with an explicit cost-sensitive treatment of missed versus unnecessary escalation.
\end{abstract}


\section{Introduction}
Healthcare is one of the clearest domains in which the distinction between a fluent language model and a useful language agent becomes operationally important. Clinical and care workflows are multi-step, stateful, and high stakes. A system may need to summarize a chart, retrieve guideline evidence, maintain dialogue context, defer when information is missing, and support a human handoff when uncertainty increases. In such settings, strong one-turn performance is not sufficient evidence of practical usefulness.

This gap matters because the healthcare AI literature is moving quickly toward agentic framing. Recent work discusses planning, tool use, retrieval, memory, and multi-step interaction as central features of LLM systems in medicine and care \cite{algaradi2025,liu2024survey,luo2025,plaat2025}. At the same time, several surveys and perspective papers warn that generic benchmark gains can be mistaken for clinical readiness \cite{chen2024eval,aljohani2025,shah2023,wornow2023,omar2024}. The main failure is not always incorrect text in isolation. Often it is a failure to preserve state, retrieve the right evidence, surface uncertainty, or stay within workflow scope.

This short paper takes a narrow position: a useful healthcare-agent benchmark needs an operational layer between static prompt-response tests and prospective clinical workflow studies. Static benchmarks remain useful, but they should be treated as one layer in a broader evidence stack.

Our contribution is threefold. First, we distinguish claims supported by evidence at the model, agent, and simulated-workflow levels. Second, we specify a reproducible episode protocol, including reference decisions, annotation units, and aggregation. Third, we provide four task templates and a cost-sensitive escalation score. The contribution is a design and measurement protocol, not a released dataset or clinical validation study.

\section{Why Healthcare Agent Evaluation Is Different}
Healthcare work is sequential rather than atomic. Documentation, triage, patient follow-up, discharge communication, and evidence-assisted decision support all unfold over time. The user goal may change mid-task. New information may arrive after an initial output. A system may need to ask a clarifying question rather than answer immediately. These features make workflow quality at least as important as local answer quality.

This point is especially visible for agents. A standalone model can appear strong if it gives a fluent answer to a static question. A healthcare agent, by contrast, may fail even when its language is polished. It can omit a contraindication, carry forward stale chart information, or retrieve irrelevant evidence while sounding persuasive. In other words, healthcare failures are often state or process failures rather than pure generation failures. This is why recent work on agent memory, retrieval-grounded generation, and dynamic medical evaluation is directly relevant to healthcare AI assessment \cite{du2026,yu2026,yan2025,yang2025rag,ceresa2025}.

The practical implication is straightforward: evaluation should track whether an agent preserves relevant context, uses evidence appropriately, recovers from interruption, and escalates when scope boundaries are reached. Recent work on realistic static and live evaluation of agent routing further motivates treating routing as an explicit behavior rather than an incidental text output \cite{yang2026twinrouterbench}. A benchmark that ignores these behaviors may be useful for capability screening while still being weak as evidence for deployment.

\section{Limits of Static Benchmarks}
Static medical QA benchmarks remain useful for testing factual recall, explanation, multilingual coverage, and specialty breadth \cite{cai2023,alonso2024,kim2024}. Hallucination benchmarks similarly help characterize unsupported claims and safety-sensitive failure modes \cite{agarwal2024,pandit2025}. However, these tasks freeze interaction into a narrow slice. They rarely test whether the model should have deferred, queried for missing context, or tracked prior state across turns.

Clinical phenotype concept recognition is another useful capability target \cite{tao2025autopcr}, but strong performance on a prompting-based extraction task likewise does not establish that an agent can manage state changes, evidence constraints, or escalation within a workflow.

This limitation becomes more serious once systems are framed as assistants embedded in care workflows. Recent benchmark directions begin to address the problem. MedAgentBench evaluates agentic tasks in EHR-like environments rather than only text prompts \cite{jiang2025}. PhysicianBench and standardized-patient-style evaluations move further toward long-horizon clinical behavior \cite{liang2026,liu2026physician,xie2026}. EHR-grounded and consultation-centered benchmarks also improve realism by anchoring evaluation in records, physician chats, and evolving patient trajectories \cite{kweon2024,wu2025bridge,gong2026,soskin2026}. Medical summarization and RAG-focused evaluations add another important axis by testing provenance retention, evidence selection, and grounded generation \cite{vanveen2023,ceresa2025,zhao2025}.

The field is therefore already moving in the right direction. The remaining challenge is conceptual integration. Different benchmarks capture different layers of useful behavior, but they are often reported side by side without a common evaluation logic. This can obscure what kind of claim a strong score actually supports and makes reusable benchmark design harder.

\section{A Three-Level Evidence Protocol}
We organize healthcare NLP agent assessment according to three evidence levels. The distinction is not a new claim that evaluation is layered; its added value is a reporting rule: every result must identify its evidence level and must not be promoted to a stronger deployment claim.

Table~\ref{tab:framework} summarizes the framework in a compact form. The table is intentionally practical: each level is tied to a deployment question, a typical metric family, and the main blind spot if that level is used alone.

{\small
\begin{table}[t]
\caption{Three-level framework for healthcare NLP agent evaluation.}
\label{tab:framework}
\centering
\setlength{\tabcolsep}{3pt}
\begin{tabular}{p{0.16\linewidth}p{0.25\linewidth}p{0.27\linewidth}p{0.22\linewidth}}
\toprule
\textbf{Level} & \textbf{Main question} & \textbf{Representative metrics} & \textbf{What it misses if used alone} \\
\midrule
Model & Does the system know relevant medicine and avoid obvious unsafe errors? & Factuality, calibration, robustness, harmful error rate, multilingual coverage & Longitudinal state, tool use, handoff quality, interruption recovery \\
Agent & Can the system execute a multi-step task with retrieval, memory, and bounded planning? & Retrieval fidelity, tool correctness, memory consistency, update behavior, abstention quality & Real workload effects, review burden, institutional constraints \\
Simulated workflow & Does the system route, hand off, and recover correctly in a designed episode? & Escalation utility, handoff completeness, state recovery, reviewable output burden & Clinical outcomes, time saved, institutional adoption \\
\bottomrule
\end{tabular}
\end{table}
}

\subsection{Model Level}
The first level measures core model behavior: factuality, calibration, fairness, robustness, and harmful error patterns. This is where static QA, explanation, and hallucination benchmarks remain appropriate. They answer questions such as whether the model knows relevant medicine, whether it makes unsupported claims, and whether it degrades under distributional or linguistic variation \cite{chen2024eval,aljohani2025,agarwal2024}.

\subsection{Agent Level}
The second level measures agent behavior during multi-step task execution. Relevant constructs include planning quality, tool-use correctness, retrieval fidelity, memory consistency, update behavior, and uncertainty handling. This is where general LLM-agent and memory research becomes useful to healthcare because it provides a vocabulary for testing write, retrieve, update, forget, and reflect operations rather than only end responses \cite{luo2025,plaat2025,du2026,yu2026}. Healthcare-specific retrieval and EHR-like benchmarks fit naturally at this level \cite{jiang2025,yang2025rag,zhao2025}.

\subsection{Simulated Workflow Level}
The third level evaluates routing, interruption recovery, handoff completeness, and the reviewability of outputs in a controlled episode. It is deliberately not a measure of time saved, clinician workload, patient safety, or care outcomes. Those claims require prospective workflow studies with local governance and human participants. A system may look strong at the model and agent levels but still fail a simulated episode if it creates an incomplete handoff, ignores a state update, or treats a routing decision as ordinary text generation \cite{wornow2023,omar2024}.

The protocol reports a vector, not a single leaderboard score: model competence, agent execution, and simulated-workflow performance. Strong results at the last level support only claims about the defined episodes. They do not establish clinical integration value.

\section{Task-Class Differences}
Documentation support, evidence retrieval, patient messaging, and triage do not fail in the same way. A benchmark package that is adequate for one task may be poorly matched to another. Table~\ref{tab:tasks} restores this task-sensitive view: a universal healthcare-agent leaderboard can obscure the changing balance between correctness, grounding, escalation, and continuity.

{\small
\begin{table}[t]
\caption{Task classes have distinct primary risks and evaluation emphasis.}
\label{tab:tasks}
\centering
\setlength{\tabcolsep}{3pt}
\begin{tabular}{p{0.24\linewidth}p{0.27\linewidth}p{0.36\linewidth}}
\toprule
\textbf{Task class} & \textbf{Primary risk} & \textbf{Evaluation emphasis} \\
\midrule
Documentation support & Omission, distortion, propagation & Update fidelity, provenance, correction review \\
Evidence retrieval and decision support & Unsupported inference, misplaced confidence & Retrieval quality, attribution, abstention \\
Patient messaging & Over-reassurance, unsafe advice, scope drift & Escalation, safe communication, uncertainty marking \\
Triage and coordination & Missed escalation, dropped state, broken handoff & Priority, interruption recovery, handoff completeness \\
\bottomrule
\end{tabular}
\end{table}
}

For example, note transformation can be locally accurate yet still be weak for triage, where the dominant risk shifts from factuality to escalation and continuity. Likewise, evidence retrieval does not establish safety for patient-facing communication without testing boundaries and harmful-advice mitigation.

\section{Episode Specification and Scoring}
The evaluation unit is a short, artifact-grounded workflow episode rather than a single prompt. Each episode has five required fields: (i) an initial context packet, (ii) a state-changing event or interruption, (iii) an allowed action space, (iv) a reference decision profile, and (v) an adjudication rubric. The context uses de-identified or synthetic artifacts; the state change identifies facts that must be retained, revised, or discarded. The action space makes clear whether the agent may answer, retrieve, revise, ask a question, abstain, or escalate. This fixed schema makes episode construction auditable and prevents post-hoc scoring criteria.

Two qualified annotators independently produce reference decisions using an episode guide. They label required facts, prohibited claims, acceptable evidence sources, the appropriate action set, and handoff fields. A third reviewer adjudicates disagreements, and a benchmark release should retain both the final label and disagreement type. This supports reproducibility without claiming that a single reference response is uniquely correct.

{\small
\begin{table}[t]
\caption{Episode templates and pre-specified primary scoring targets.}
\label{tab:proposal}
\centering
\setlength{\tabcolsep}{3pt}
\begin{tabular}{p{0.21\linewidth}p{0.22\linewidth}p{0.24\linewidth}p{0.20\linewidth}}
\toprule
\textbf{Episode} & \textbf{State-changing input} & \textbf{Reference decision} & \textbf{Primary score} \\
\midrule
Documentation update & Note draft plus late chart detail & Revise facts; preserve unaffected content & update fidelity \\
Evidence retrieval & Query plus evidence set and new constraint & Retrieve admissible sources; attribute claims & grounded-decision score \\
Patient messaging & Portal message plus risk cue & Reply within scope or escalate & escalation utility \\
Triage handoff & Longitudinal thread plus interruption & Preserve state; route with required fields & handoff completeness \\
\bottomrule
\end{tabular}
\end{table}
}

For every episode, scores are calculated from atomic annotations rather than global impressions. State continuity is scored as the fraction of required facts retained or correctly revised, minus contradictions and stale facts. Evidence traceability is scored as the proportion of substantive claims linked to an admissible source, with penalties for unsupported or mismatched citations. Handoff completeness is scored as the fraction of required routing fields that are present and correct. All score components and weights are reported by task class.

Escalation is evaluated as a decision, not as a generic safety preference. Let $E$ denote the reference escalation label and $\hat{E}$ the agent action. We report sensitivity, specificity, and a configurable utility, $U = u_{TP}TP + u_{TN}TN - c_{FN}FN - c_{FP}FP$, where $c_{FN} > c_{FP}$ when missed escalation is judged more harmful than unnecessary escalation. The cost matrix must be declared before evaluation and reviewed by the task's clinical governance group. This makes the trade-off visible instead of silently rewarding either excessive deferral or unsafe completion.

\section{Illustrative Failure Analysis}
Consider a patient message reporting worsening shortness of breath after a recent medication change. A static benchmark may reward a correct explanation or summary, but it can miss the clinically salient failure: producing a polished reply instead of initiating escalation. The episode protocol instead asks whether warning signals are recognized, prior context is retained, and the handoff contains the required information. In this setting, a brief acknowledgment plus escalation can be preferable to a detailed answer.

\section{Design Requirements and Boundaries}
The protocol operationalizes four requirements and sets a clear boundary on its claims.

\subsection{State Continuity}
Episodes test whether relevant information is preserved, refreshed, and discarded across turns. The annotation guide distinguishes retained facts, revised facts, and facts that must no longer be used. Memory is therefore evaluated as a governance problem, not only as a context-window problem \cite{du2026,yan2025,liu2026memory}.

\subsection{Evidence Traceability}
The rubric records admissible sources and requires claim-level links to them. This permits separate scoring of retrieval, attribution, and unsupported inference, rather than treating citations as a cosmetic feature \cite{yang2025rag,ceresa2025,zhao2025}. Work on plan-guided retrieval and execution provenance in LLM agents complements this approach \cite{tao2026grasp,AgentTraces}.

\subsection{Escalation Sensitivity}
Episodes include an explicit appropriate-action set and a pre-declared cost matrix. This allows safe abstention and escalation to be scored alongside unnecessary escalation, particularly for triage and patient-facing communication \cite{aljohani2025,agarwal2024}.

\subsection{Institutional Realism}
Episodes can represent fragmented records, missing information, interruptions, and specialty-specific conventions. They do not reproduce institutional implementation or patient outcomes. Such claims require prospective evaluation; the present protocol is intended to make later studies more targeted.

\section{Conclusion}
Healthcare NLP agents should be evaluated as workflow-bound systems rather than as isolated response generators. Static medical benchmarks remain useful, but they do not justify claims about deployable assistance by themselves. We provide an episode-level protocol that separates evidence levels and makes state continuity, traceability, escalation, and handoff behavior reproducibly scorable.

The immediate next step is a small, clinically governed pilot that releases episode guides, adjudication decisions, score distributions, and failure analyses. Until then, this paper should be read as a testable measurement protocol rather than evidence that any agent improves healthcare work.

\balance
\bibliographystyle{IEEEtran}
\bibliography{healthcare_nlp_agents_review}

\end{document}